\documentclass{article} 
\PassOptionsToPackage{table}{xcolor}
\usepackage{fpd_arxiv,times}

\usepackage{amsmath,amsfonts,bm}

\def\eqref#1{equation~\ref{#1}}

\def\1{\bm{1}}

\DeclareMathAlphabet{\mathsfit}{\encodingdefault}{\sfdefault}{m}{sl}
\SetMathAlphabet{\mathsfit}{bold}{\encodingdefault}{\sfdefault}{bx}{n}

\usepackage{url}
\usepackage{graphicx}
\usepackage{booktabs}
\usepackage{amsmath,amssymb}
\usepackage{multirow}
\usepackage{xcolor}
\usepackage{algorithm}
\usepackage{algorithmic}
\usepackage{appendix}

\usepackage{pifont}
\usepackage{hyperref}
\newcommand{\cmark}{\ding{51}}
\newcommand{\xmark}{\ding{55}}

\newcommand{\methodname}{FPD-Drive}

\newsavebox{\fpdtablemeasurebox}
\newcommand{\fpdfullwidthtabular}[3]{%
  \sbox{\fpdtablemeasurebox}{\begin{tabular}{#2}#3\end{tabular}}%
  \setlength{\tabcolsep}{\dimexpr\tabcolsep+(\linewidth-\wd\fpdtablemeasurebox)/#1\relax}%
  \begin{tabular}{#2}#3\end{tabular}%
}

\title{%
  \methodname: Future Policy Distillation for\\
  End-to-End Autonomous Driving%
}

\author{%
  Li-Heng Chen\textsuperscript{1,2,*}\quad
  Qifeng Chen\textsuperscript{1,*,\ensuremath{\dagger}}\quad
  Chengwei Wu\textsuperscript{1}\quad
  Yingjie Zhang\textsuperscript{1}\\[0.4em]
  Sheng Yang\textsuperscript{1}\quad
  Hongbo Fu\textsuperscript{2,\ensuremath{\ddagger}}\quad
  Shaoqing Ren\textsuperscript{1,3,\ensuremath{\ddagger}}\\[0.9em]
  {\small\textsuperscript{1}\,NIO}\\[0.2em]
  {\small\textsuperscript{2}\,The Hong Kong University of Science and Technology}\\[0.2em]
  {\small\textsuperscript{3}\,University of Science and Technology of China}\\[0.7em]
  {\small\textsuperscript{*}\,Equal contribution.\quad
   \textsuperscript{\ensuremath{\dagger}}\,Project lead.\quad
   \textsuperscript{\ensuremath{\ddagger}}\,Corresponding authors.}%
}
\date{}

\hypersetup{%
  pdftitle={FPD-Drive: Future Policy Distillation for End-to-End Autonomous Driving},
  pdfauthor={Li-Heng Chen, Qifeng Chen, Chengwei Wu, Yingjie Zhang, Sheng Yang, Hongbo Fu, Shaoqing Ren},
  pdfsubject={End-to-end autonomous driving},
  pdfkeywords={autonomous driving, policy distillation, trajectory planning},
  hidelinks
}

\begin{document}

\maketitle

\begin{abstract}
End-to-end autonomous driving models are typically trained on real-world driving logs, which record only the trajectory actually executed in each scene. This provides limited supervision for learning alternative valid behaviors.
Multi-target supervision can mitigate this limitation, but broader trajectory coverage makes safe candidate selection more challenging.
We propose \methodname{}, an end-to-end planner that distills privileged future knowledge into a visual driving policy to improve both multi-trajectory supervision and safety selection.
Specifically, we introduce Future-Privileged Multi-Target Distillation to transfer diverse, high-quality trajectory targets from a teacher with access to recorded future observations to the student.
These targets extend supervision beyond the single recorded human trajectory.
For candidate selection, we further introduce Safety-Aware Score Correction to correct ranking errors between nearby safe and unsafe candidates and raise underestimated safety scores of safe candidates.
At deployment, the student requires neither future observations nor teacher inference.
Our base model achieves 94.9 PDMS on NAVSIM v1 \textit{navtest} and 55.9 EPDMS on NAVSIM v2 \textit{navhard}.
The scaled \methodname{}-Pro improves these scores to 95.5 PDMS and 58.1 EPDMS, respectively, achieving SOTA performance among the compared methods on both benchmarks.
Zero-shot closed-loop evaluation on HUGSIM further yields an HD-Score of 45.1 and RC of 55.3.  
Project page: \url{https://countnemochan.github.io/FPD-Drive/}.

\end{abstract}

\section{Introduction}
\label{sec:intro}

End-to-end autonomous driving aims to learn planning policies directly from sensor observations~\citep{hu2023uniad,jiang2023vad}.
Advances in visual representation learning and generative trajectory modeling have enabled planners to produce and evaluate multiple candidate trajectories for a driving scene~\citep{kirby2026drivor,liao2025diffusiondrive,xing2025goalflow}.
Effective planning requires both generating valid alternatives and reliably selecting a safe trajectory for execution.
However, learning these capabilities from human drivers is challenging, because each logged driving clip used for training typically provides only one executed trajectory and its associated visual observations.
As a result, it provides limited guidance for learning and evaluating alternative trajectories.

A recorded human trajectory represents one choice among several potentially acceptable driving behaviors.
For example, different deceleration timings or speed profiles may be valid in the same scene, although only one is recorded.
Closeness to the demonstration therefore cannot fully determine an alternative trajectory's quality.
To supervise such alternatives, recent methods use predefined trajectory vocabularies or teachers pretrained on rule-generated targets~\citep{wang2026drivejepa,ang2026clover,chen2026curiousvla}.
These approaches broaden supervision, but constructing alternatives that are both diverse and scene-appropriate remains challenging.
Predefined vocabularies and handcrafted generation rules can restrict the range of candidate behaviors covered by supervision. And recorded future observations reveal how the scene's structure and evolution can help guide planning alternative trajectories.
These motivates a privileged teacher that uses these observations to generate additional trajectory targets.

To benefit from broader trajectory coverage, the planner should also learn to select a safe candidate reliably.
Planning metrics provide supervision beyond imitation to help select candidates, but accurate safety assessment remains challenging~\citep{li2024hydramdp,kirby2026drivor}.
Geometrically similar trajectories can differ sharply in safety metrics, and a scorer may favor an unsafe candidate over a nearby safe alternative.
Safe candidates may also be overlooked if the scorer underestimates their safety.
These two types of scoring errors motivate targeted score correction to distinguish nearby safe and unsafe candidates and raise the scores of underestimated safe alternatives.

We propose \methodname{}, a future policy distillation framework that enriches trajectory supervision with privileged future knowledge while jointly improving candidate selection.
The teacher and student share a coarse-to-fine visual planner that refines trajectory proposals through residual diffusion.
During training, the teacher uses recorded future observations to guide trajectory refinement and scoring. We then evaluate its candidates using planning metrics and select high-quality, geometrically diverse trajectory targets.
\emph{Future-Privileged Multi-Target Distillation} transfers these targets to the student, extending supervision beyond the recorded demonstration.
For candidate selection, \emph{Safety-Aware Score Correction} uses evaluated safety labels for the student's own trajectories to correct scoring errors.
It penalizes highly ranked unsafe candidates relative to nearby safe alternatives and increases confidence in underestimated safe candidates.

The deployed student requires neither future observations nor teacher inference.
\methodname{} achieves state-of-the-art performance on NAVSIM v1 and v2.
Zero-shot evaluation on HUGSIM further supports its transfer to closed-loop driving. Our contributions are threefold:


\begin{itemize}
    \item We propose \methodname{}, which uses recorded future observations to help a teacher generate additional high-quality trajectories for distilling a visual student. Only the student is needed at deployment.
    \item We introduce \emph{Future-Privileged Multi-Target Distillation}, which transfers diverse, high-quality trajectories from a future-conditioned teacher to the student, extending supervision beyond the single recorded demonstration.
    \item We develop \emph{Safety-Aware Score Correction} to correct the relative safety scores of unsafe--safe pairs and raise underestimated safety scores for safe trajectories.
\end{itemize}

\section{Related Work}
\label{sec:related_work}

\paragraph{End-to-end autonomous driving.}
End-to-end autonomous driving learns planning policies directly from sensor observations.
Research on planning representations includes multimodal sensor fusion~\citep{prakash2021transfuser} and integrating perception, prediction, and planning~\citep{hu2023uniad,jiang2023vad}.
To account for multiple acceptable driving behaviors, probabilistic and generative planners model distributions over possible trajectories~\citep{jiang2026vadv2,liao2025diffusiondrive,xing2025goalflow}.
ResAD~\citep{zheng2025resad} further improves trajectory representations through
normalized residual modeling to address imbalanced
optimization~\citep{zheng2025resad}.
Beyond generation, metric-based supervision and generalized trajectory evaluation address reliable selection among alternative behaviors~\citep{li2024hydramdp,li2025gtrs,kirby2026drivor}.
World models and world-action models (WAMs) introduce predictive information by learning how driving scenes evolve.
Visual forecasting~\citep{wang2024drivewm,gao2024vista} and latent prediction~\citep{li2025law,wang2026latentwam} represent future scene dynamics at different levels.
Predicted future states support candidate evaluation~\citep{li2025wote,zheng2025world4drive,zhong2026dawam}, while joint world--action modeling couples future scene prediction with trajectory generation~\citep{zhang2025epona,liu2026mmfuture,wang2026wajepa}.
Vision-language-action (VLA) models draw on pretrained semantic knowledge and language understanding to support planning~\citep{xu2023drivegpt4,shao2024lmdrive,hwang2024emma}.
These methods connect semantic scene understanding with driving actions by conditioning trajectory generation on visual-language representations~\citep{zhou2026opendrivevla,li2026recogdrive,wang2026chainflowvla}.
Graph-based visual question answering connects scene understanding with driving decisions~\citep{sima2024drivelm}, while reasoning-augmented policies integrate semantic reasoning with action generation~\citep{zhou2025autovla,nvidia2025alpamayo}.
To balance planning performance and inference latency, \methodname{} adopts a coarse-to-fine visual planner as the common architecture for the teacher and student. It extends DrivoR with a DiT module that refines the regressed trajectories.~\citep{kirby2026drivor}.
Furthermore, recorded future observations enrich trajectory supervision through a privileged teacher during training.

\paragraph{Distillation for autonomous driving.}
Knowledge distillation and privileged learning have attracted growing interest in autonomous driving.
Teachers with access to ground-truth road structure and traffic-participant states provide trajectory targets for sensor-based students~\citep{chen2020lbc,zhang2021roach,xiaomi2026drivezero}.
Feature distillation also transfers scene and planning representations from these teachers~\citep{zhang2023cat,jia2023driveadapter,yu2025distilldrive}.
Recent VLA methods condition teachers on privileged future information and distill token distributions into student driving policies~\citep{cao2026evodrivevla,dou2026firevla}.
To broaden trajectory, Curious-VLA~\citep{chen2026curiousvla} expands training data with trajectories generated by the ReCogDrive-VLA~\cite{li2026recogdrive}.
Some concurrent methods explore multi-target trajectory supervision using predefined trajectory vocabularies~\citep{wang2026drivejepa, sun2026sparsedrivev2} or teacher models pretrained on rule-generated trajectories~\citep{ang2026clover}. 
Without relying on predefined trajectories or vocabulary,
\methodname{} combines a future-conditioned teacher with multi-target trajectory policy to distill a student. We further introduce Safety-Aware Score Correction to improve
safety assessment and candidate ranking.

\section{Method}
\label{sec:method}

\begin{figure*}[t]
    \centering
    \includegraphics[width=1.0\linewidth]{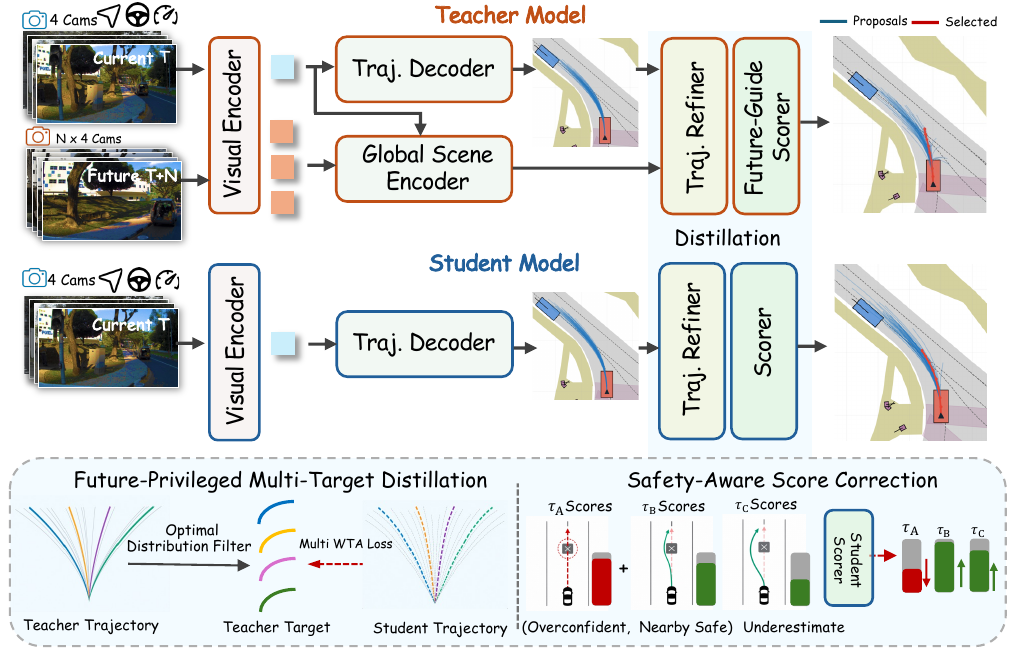}
    \caption{\textbf{Overview of \methodname{}.}
    The teacher--student framework employs a common coarse-to-fine architecture comprising a visual encoder, a proposal decoder, a diffusion refiner, and a scorer (Sec.~\ref{sec:method:planner}).
    Recorded future observations additionally guide the teacher's refinement and scoring (Sec.~\ref{sec:method:teacher}).
    \textbf{Future-Privileged Multi-Target Distillation} transfers high-quality, geometrically diverse trajectories selected from evaluated teacher candidates to supervise the student's proposal decoder and refiner (Sec.~\ref{sec:method:multitarget}).
    \textbf{Safety-Aware Score Correction} uses evaluated student candidates to penalize highly ranked unsafe trajectories relative to nearby safe alternatives and increase confidence in underestimated safe candidates (Sec.~\ref{sec:method:safety}).
    Only the student is used at deployment.}
    \label{fig:overview}
\end{figure*}

\paragraph{Overview.}

Fig.~\ref{fig:overview} illustrates the overall architecture of \methodname{}.
At planning time $t$, the input $o_t$ consists of four current camera views, the current ego-state, and a navigation command.
The planner generates $M=64$ candidate trajectories and selects one for execution.
Each trajectory contains $H=8$ future poses, represented as $\tau=\{(x_h,y_h,\psi_h)\}_{h=1}^{H}$ in the ego coordinate frame at time $t$.
Here, $(x_h,y_h)$ and $\psi_h$ denote position and heading, respectively.
Training clips additionally provide a demonstrated trajectory $\tau^{\mathrm{gt}}$ and recorded future camera observations $\mathcal{F}_t$.
The future observations are used as privileged teacher inputs and are unavailable to the student at deployment.

\subsection{Coarse-to-Fine Trajectory Modeling}
\label{sec:method:planner}

We separate trajectory generation into scene-conditioned proposals and residual refinement.
As illustrated in Fig.~\ref{fig:overview}, the proposal decoder provides a finite set of starting trajectories, which may cluster within a limited region of trajectory space.
A diffusion refiner then models continuous adjustments around these proposals to improve trajectory quality and broaden candidate coverage.
This provides more alternatives for selection under different driving preferences.
We adopt this coarse-to-fine structure as the common backbone for the teacher and student.

\paragraph{Trajectory proposals.}
Following DrivoR~\citep{kirby2026drivor}, both the teacher and student encode current multi-view images into scene register tokens.
We initialize $M$ trajectory tokens by adding the ego-state and navigation embedding $\mathbf{e}$ to learned queries $\{\mathbf{q}_m\}_{m=1}^{M}$.
A transformer decoder updates these tokens using the scene tokens $\mathbf{h}$ and predicts the proposals:
{\small
\begin{equation}
    \mathcal{B}=\{b_m\}_{m=1}^{M}
    =D\!\left(\{\mathbf{q}_m+\mathbf{e}\}_{m=1}^{M},\mathbf{h}\right),
    \qquad b_m\in\mathbb{R}^{H\times3}.
    \label{eq:method:proposals}
\end{equation}
}

\paragraph{Residual diffusion.}
Drawing on residual trajectory modeling in ResAD~\citep{zheng2025resad} and proposal-based diffusion refinement in ChainFlow-VLA~\citep{wang2026chainflowvla}, we use a Diffusion Transformer (DiT) to predict residual corrections to coarse trajectories.
For a proposal $b_m$ and its training target $\tilde y_m$, we define the residual $r_m=\tilde y_m-\operatorname{sg}(b_m)$, where $\operatorname{sg}$ denotes the stop-gradient operator.
At a sampled diffusion step $k$, we construct:
{\small
\begin{equation}
    z_{k,m}=\sqrt{\bar\alpha_k}\,r_m
    +\sqrt{1-\bar\alpha_k}\,\epsilon_m,
    \qquad \epsilon_m\sim\mathcal{N}(0,\mathbf{I}),
    \label{eq:method:corruption}
\end{equation}
}\ignorespaces
where $\bar\alpha_k$ is the cumulative noise schedule.
A DiT predicts the noise as $\hat\epsilon_m=\epsilon_\theta(z_{k,m},k;\mathbf{h},\mathbf{e})$.
Its blocks apply self-attention over trajectory poses and cross-attention to scene tokens.
The refined trajectory $\hat\tau_m$ is reconstructed from proposal $b_m$ as follows, and will be send to the scorer.
{\small
\begin{equation}
    \hat\tau_m=\operatorname{sg}(b_m)+
    \frac{z_{k,m}-\sqrt{1-\bar\alpha_k}\,\hat\epsilon_m}
         {\sqrt{\bar\alpha_k}}.
    \label{eq:method:reconstruction}
\end{equation}
}\ignorespaces

\paragraph{Candidate scoring.}
We use a transformer-based trajectory scorer with self-attention among candidates and cross-attention to scene tokens.
Separate MLP prediction heads estimate the five components of the Predictive Driver Model Score (PDMS).
The predicted metric components are aggregated into a selection score $\hat Q_{\mathbf{w}}(\tau)$~\citep{dauner2024navsim}.
We select the candidate with the highest score for execution.
The aggregation weights $\mathbf{w}$ can be adjusted at inference to reflect different driving preferences without retraining~\citep{kirby2026drivor}.

\subsection{Learning a Future-Privileged Teacher}
\label{sec:method:teacher}

The trajectory generator can produce multiple candidate distributions, but a single demonstration provides limited supervision for these alternatives.
We therefore train a teacher model using recorded future observations as privileged conditioning.
The teacher aggregates observations across multiple future time steps into a global scene representation.
This representation guides trajectory refinement to produce additional high-quality target behaviors.

\paragraph{Future-conditioned planning.}
We reuse the visual encoder to extract visual tokens from recorded future images.
A trainable global scene encoder $F$ combines these tokens with the current scene tokens $\mathbf{h}$ to form a global scene representation $\mathbf{f}$:
{\small
\begin{equation}
    \mathbf{f}=F\!\left(\mathbf{h},\bar E(\mathcal{F}_t))\right.
    \label{eq:method:future}
\end{equation}
}\ignorespaces
where $\bar E$ denotes the visual encoder used for future observations.
The global scene encoder initializes queries from $\mathbf{h}$ and combines current and future information through spatial, temporal, and cross-attention.
The global scene representation $\mathbf{f}$ is shared across candidates and conditions, both the teacher's refiner and scorer, through cross-attention.

\paragraph{Teacher training and target generation.}
We train the teacher using recorded trajectories and metric supervision.
Although each scene provides only a single trajectory, future observations provide additional scene context that helps the teacher generate higher-quality alternatives.
After training, we freeze the teacher and sample a bank $\mathcal{T}=\{\tau_j^T\}_{j=1}^{M}$ for each scene.
A PDM-closed metric evaluator assigns component labels $m_r(\tau)$ to each candidate.
These evaluated labels determine target eligibility, allowing trajectories that differ from the human driver to be retained.
We then select high-quality, geometrically distinct candidates to supervise the student.

\subsection{Future-Privileged Multi-Target Distillation}
\label{sec:method:multitarget}

Transferring only the teacher's highest-scoring trajectory would retain the single-target limitation.
We instead select a small set of high-quality alternatives and assign each to a different student proposal.
This separates target quality, target diversity, and supervision assignment.

\paragraph{Selecting teacher targets.}
We retain valid candidates that pass the safety check of PDMS.
We initialize the target set with the highest-scoring trajectory and iteratively expand it by adding the candidate that maximizes the minimum average displacement error (ADE) to the selected trajectories.
Selected targets must have a pairwise ADE of at least $1$\,m.
The resulting target set $\mathcal{Y}=\{y_j\}_{j=1}^{J}$ contains $J\leq K$ trajectories, where $K=4$ is the maximum number of targets per scene.

\paragraph{Target assignment and residual construction.}
We use optimal bipartite matching to assign the $J$ teacher targets $\mathcal{Y}$ to distinct proposals from the full student set $\mathcal{B}$.
The assignment minimizes the total ADE:
{\small
\begin{equation}
    \pi^*=\arg\min_{\pi\ \mathrm{injective}}
    \sum_{j=1}^{J}d_{xy}(b_{\pi(j)},y_j).
    \label{eq:method:matching}
\end{equation}
}\ignorespaces
Here, $d_{xy}(a,b)$ is the mean Euclidean distance between corresponding positions, and $\pi^*(j)$ indexes the student proposal matched to teacher target $y_j$.
This yields $J$ matched pairs for trajectory supervision.

To further construct noisy residuals for all $M$ candidates, we assign each unmatched proposal its nearest teacher target in ADE while preserving the matched assignments.
Specifically, following Eq.~\eqref{eq:method:corruption}, we add Gaussian noise to the residual between each proposal and its assigned teacher target to form the DiT input.
Denoising reconstructs all $M$ candidates, while the following trajectory losses supervise only the $J$ matched pairs:
{\small
\begin{equation}
\begin{aligned}
    \mathcal{L}_{\mathrm{prop}}&=\frac{1}{J}\sum_{j=1}^{J}d_1(b_{\pi^*(j)},y_j), \\
    \mathcal{L}_{\mathrm{diff}}&=\mathbb{E}_{k,\epsilon}\!\left[
    \frac{1}{HJ}\sum_{j=1}^{J}\|\hat\epsilon_{\pi^*(j)}-\epsilon_{\pi^*(j)}\|_F^2\right],\\
    \mathcal{L}_{\mathrm{traj}}&=\mathbb{E}_{k,\epsilon}\!\left[
    \frac{1}{J}\sum_{j=1}^{J}d_1(\hat\tau_{\pi^*(j)},y_j)\right].
\end{aligned}
    \label{eq:method:trajectory-losses}
\end{equation}
}\ignorespaces
Here, $d_1(a,b)$ denotes the mean per-pose sum of absolute position and heading errors.
$\mathbb{E}_{k,\epsilon}$ denotes the expectation $\mathbb{E}$ over the sampled diffusion step $k$ and Gaussian noise $\epsilon$.
$\mathcal{L}_{\mathrm{prop}}$ supervises the proposal decoder with an $L_1$ trajectory loss against matched teacher targets.
The diffusion refiner is trained with $\mathcal{L}_{\mathrm{diff}}$, a mean squared error loss on predicted noise, and $\mathcal{L}_{\mathrm{traj}}$, an $L_1$ trajectory loss between refined candidates and the same matched targets.

\subsection{Safety-Aware Score Correction}
\label{sec:method:safety}

Multi-target distillation provides diverse trajectory supervision, but safe candidate selection also requires accurate safety scores.
Highly ranked unsafe candidates can displace safe alternatives, while underestimated safe candidates may be overlooked. In performance evaluation, a zero safety label reduces the overall score to zero, regardless of other metric values.
We therefore introduce Safety-Aware Score Correction, using evaluated safety labels for the student's own candidates to correct both errors.

\paragraph{Score supervision.}
The scorer is supervised on all $M$ student trajectories using binary cross-entropy:
{\small
\begin{equation}
    \mathcal{L}_{\mathrm{metric}}=
    \frac{1}{M}\sum_{m=1}^{M}\sum_{r\in\mathcal{R}}
    \operatorname{BCE}(\hat p_r(\hat\tau_m),\tilde m_r(\hat\tau_m)).
    \label{eq:method:metric}
\end{equation}
}\ignorespaces
Here, $\hat p_r(\tau)$ and $\tilde m_r(\tau)$ denote the predicted score and corresponding training label for component $r$, respectively.
$\mathcal{R}$ indexes the five metric components. The loss is averaged over the $M$ student trajectories and summed over these components.

\paragraph{Correcting highly ranked unsafe candidates.}
We penalize candidates with high predicted selection scores that fail the evaluated safety checks.
We pair each with the nearest eligible alternative in ADE that passes the safety checks within the same scene.
Pairs must satisfy geometric proximity constraints; we skip candidates without an eligible alternative.
This strategy aims to improve the scorer's ability to distinguish safe and unsafe trajectories near the safety boundary.
A pairwise margin loss is applied to the safety logits of each unsafe--safe pair $(\tau^-,\tau^+)$:
{\small
\begin{equation}
    \mathcal{L}_{\mathrm{neg}}=
    \sum_{r\in\mathcal{S}}\frac{1}{\max(1,|\mathcal{P}_r|)}
    \sum_{(\tau^-,\tau^+)\in\mathcal{P}_r}
    \operatorname{softplus}\!\left(
    1-\ell_r(\tau^+)+\ell_r(\tau^-)\right),
    \label{eq:method:negative}
\end{equation}
}\ignorespaces
where $\mathcal{S}=\{\mathrm{NC},\mathrm{DAC}\}$ indexes the safety metric components (Appendix~\ref{app:navsim-metrics}) and $\mathcal{P}_r$ contains pairs whose unsafe candidate fails component $r$.
Here, $\ell_r$ denotes the predicted component logit.
This loss encourages lower logits for unsafe candidates and higher logits for their safe alternatives on the failed safety components.

\paragraph{Correcting underestimated safe candidates.}
Candidates that pass all safety checks may still receive low predicted safety scores.
For these candidates, we consider a safety component underestimated if its predicted probability is below 0.9.
An additional positive-label binary cross-entropy loss selectively strengthens supervision on these components:
{\small
\begin{equation}
    \mathcal{L}_{\mathrm{pos}}=
    \sum_{r\in\mathcal{S}}\frac{1}{\max(1,|\mathcal{H}_r|)}
    \sum_{\tau\in\mathcal{H}_r}\operatorname{softplus}(-\ell_r(\tau)).
    \label{eq:method:positive}
\end{equation}
}\ignorespaces
Here, $\mathcal{H}_r$ contains selected safe candidates whose predicted score for component $r$ is underestimated.
Both correction losses act on the safety logits, with candidate selection and evaluator labels detached.

\subsection{Optimization and Deployment}
\label{sec:method:optimization}

Training proceeds in three stages.
First, we train a base coarse-to-fine visual planner using recorded supervision.
Second, we initialize the teacher from this base and introduce future observations through the global scene encoder to condition refinement and scoring.
Both stages use proposal, diffusion, trajectory reconstruction, and score losses, with demonstrations as trajectory targets and evaluated metrics as score labels.
Third, we freeze the teacher and cache its selected targets.
The student is initialized from the first-stage base and fine-tuned with multi-target distillation and Safety-Aware Score Correction:
{\small
\begin{equation}
\begin{aligned}
    \mathcal{L}_{\mathrm{student}}={}&
    \lambda_{\mathrm{prop}}\mathcal{L}_{\mathrm{prop}}
    +\lambda_{\mathrm{diff}}\mathcal{L}_{\mathrm{diff}}
    +\lambda_{\mathrm{traj}}\mathcal{L}_{\mathrm{traj}}\\
    +&\lambda_{\mathrm{metric}}\mathcal{L}_{\mathrm{metric}}
    +\lambda_{\mathrm{neg}}\mathcal{L}_{\mathrm{neg}}
    +\lambda_{\mathrm{pos}}\mathcal{L}_{\mathrm{pos}}.
\end{aligned}
    \label{eq:method:total}
\end{equation}
}\ignorespaces

%

We use future observations, teacher, and metric evaluator only during training or target preparation.


\section{Experiments}
\label{sec:exp}

\begin{figure*}[ht]
    \centering
    \includegraphics[width=1.0\linewidth]{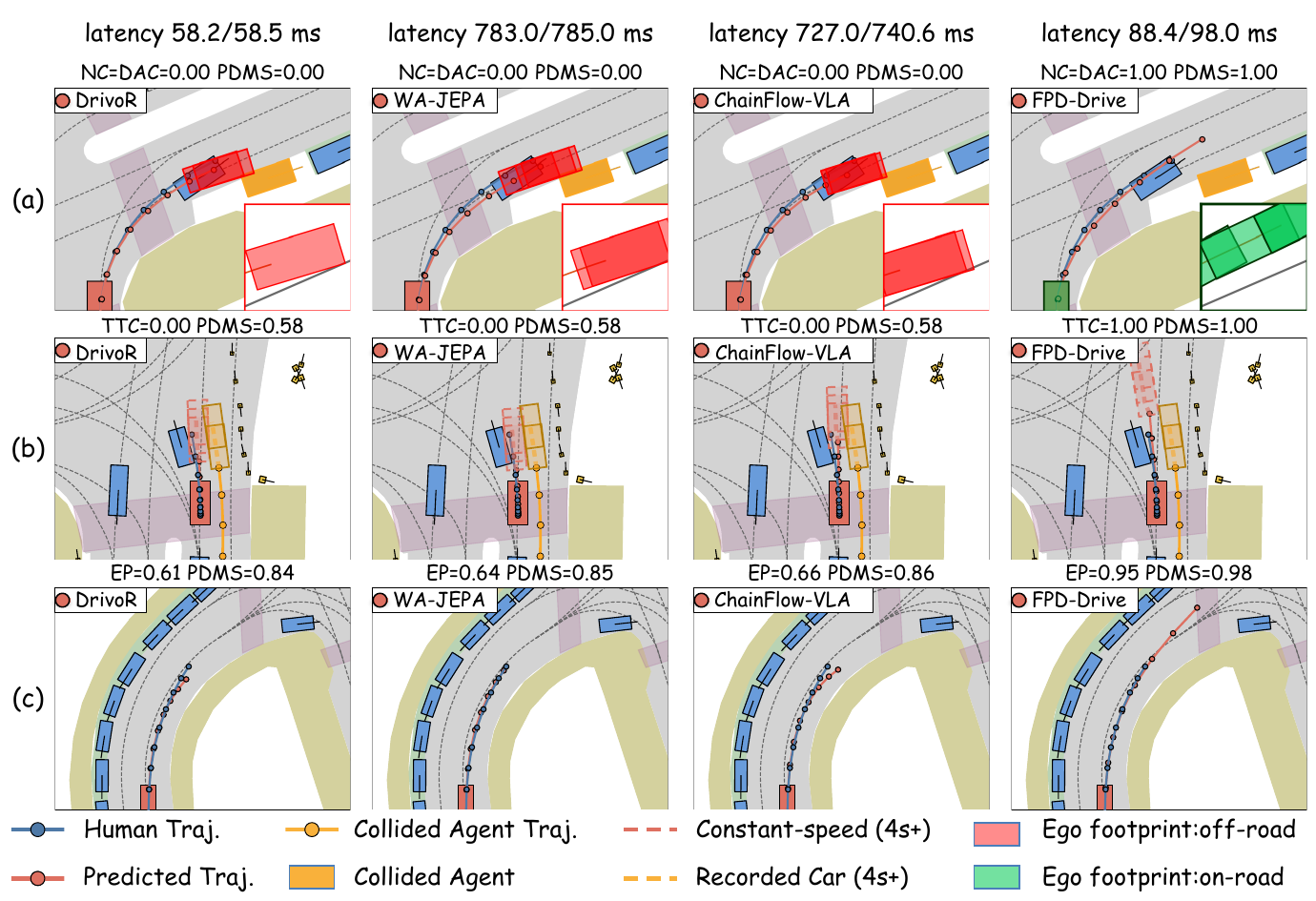}
    \caption{\textbf{Qualitative comparison on NAVSIM v1 \texttt{navtest}.} We show the metrics for each scene, with PDMS and inference latency as mean/p99 in milliseconds.}
    \label{fig:exp:qual1}
\end{figure*}

\begin{table*}[!t]
  \centering
  \small
  \caption{\textbf{Comparison on the NAVSIM v1 \texttt{navtest}}.
    Best overall PDMS is in \textbf{bold}, second-best is \underline{underlined}.}
  \label{tab:navsim-v1-leaderboard}
  \begingroup
  \setlength{\tabcolsep}{5pt}
  \renewcommand{\arraystretch}{1.08}
    \fpdfullwidthtabular{16}{lccccccc}{
      \toprule
      \multirow{2}{*}{\textbf{Entry}} &
      \multirow{2}{*}{\textbf{Venue}} &
      \multicolumn{5}{c}{\textbf{NAVSIM v1}} &
      \multirow{2}{*}{\textbf{PDMS}$\uparrow$} \\
      \cmidrule(lr){3-7}
      & & NC$\uparrow$ & DAC$\uparrow$ & EP$\uparrow$ & TTC$\uparrow$ & C$\uparrow$ & \\
      \midrule
      Human Driver & -- & 100 & 100 & 87.5 & 100 & 99.9 & 94.8 \\
      MeanFuser & CVPR 26 & 98.6 & 97.0 & 82.8 & 95.0 & 100.0 & 89.0 \\
      ReCogDrive & ICLR 26 & 97.9 & 97.3 & 87.3 & 94.9 & 100.0 & 90.8 \\
      WA-JEPA & arXiv 26.08 & 99.4 & 98.2 & 85.3 & 97.2 & 100.0 & 91.6 \\ 
      DriveFine & ECCV 26 & 98.8 & 99.2 & 86.9 & 96.2 & 100.0 & 91.8 \\
      DriveSuprim & AAAI 26 & 98.6 & 98.6 & 91.3 & 95.5 & 100.0 & 93.5 \\
      CLOVER & arXiv 26.05 & 98.6 & 99.1 & 92.8 & 95.8 & 100.0 & 94.5 \\
      Drive-JEPA & arXiv 26.01 & 99.1 & 98.2 & 91.4 & 96.2 & 99.9 & 93.7 \\
      DrivoR & CVPR 26 & 99.0 & 98.9 & 90.0 & 96.7 & 100.0 & 93.7 \\
      DrivoR-Scale & CVPR 26 & 99.1 & 99.2 & 91.6 & 96.9 & 100.0 & 94.6 \\
      ChainFlow-VLA & arXiv 26.05 & 99.1 & 99.3 & 92.0 & 97.0 & 100.0 & 94.8 \\
      DriveZero & arXiv 26.09 & 99.0 & 99.2 & 93.1 & 96.0 & 100.0 & 94.8 \\
      DriveZero-Scale & arXiv 26.09 & 99.2 & 99.4 & 92.5 & 97.3 & 99.9 & \underline{95.3} \\      
      \midrule
      \rowcolor{gray!12}
      \textbf{\methodname{}} & -- & 99.2 & 99.5 & 91.3 & 97.5 & 99.9 & 94.9 \\
      \rowcolor{gray!12}
      \textbf{\methodname{}-Pro} & -- & 99.3 & 99.3 & 93.1 & 97.4 & 100.0 & \textbf{95.5} \\
      \bottomrule
    }
  \endgroup
\end{table*}
\begin{table*}[t]
  \centering
  \small
  \caption{\textbf{Comparison on the NAVSIM v2
    \texttt{navhard}}.
    Best combined EPDMS is in \textbf{bold}, second-best is \underline{underlined}.}
  \label{tab:navsim-v2-leaderboard}
  \begingroup
  \setlength{\tabcolsep}{2.5pt}
  \renewcommand{\arraystretch}{1.08}
    \fpdfullwidthtabular{26}{lcccccccccccc}{
      \toprule
      \multirow{2}{*}{\textbf{Entry}} & \multirow{2}{*}{\textbf{Venue}}
      & \multirow{2}{*}{\textbf{Stage}} & \multicolumn{9}{c}{\textbf{NAVSIM v2}}
      & \multirow{2}{*}{\textbf{EPDMS}$\uparrow$} \\
      \cmidrule(lr){4-12}
      & & & NC$\uparrow$ & DAC$\uparrow$ & DDC$\uparrow$ & TLC$\uparrow$ & EP$\uparrow$ & TTC$\uparrow$ & LK$\uparrow$ & HC$\uparrow$ & EC$\uparrow$ & \\
      \midrule
      \multirow{2}{*}{RAP} & \multirow{2}{*}{ICLR 26} & 1 & 97.1 & 94.4 & 98.8 & 99.8 & 83.9 & 96.9 & 94.7 & 96.4 & 66.2 & \multirow{2}{*}{39.6} \\
       & & 2 & 83.2 & 83.9 & 87.4 & 98.0 & 86.9 & 80.4 & 52.3 & 95.2 & 52.4 &  \\
      \addlinespace[2pt]
      \multirow{2}{*}{ZTRS} & \multirow{2}{*}{ECCV 26} & 1 & 98.9 & 97.6 & 100.0 & 100.0 & 66.7 & 98.9 & 96.2 & 96.7 & 44.0 & \multirow{2}{*}{48.1} \\
       & & 2 & 91.1 & 90.4 & 95.8 & 99.0 & 63.6 & 89.8 & 60.4 & 97.6 & 66.1 &  \\
      \addlinespace[2pt]
      \multirow{2}{*}{GuideFlow} & \multirow{2}{*}{CVPR 26} & 1 & 99.6 & 98.0 & 99.4 & 99.3 & 79.7 & 99.3 & 94.9 & 97.1 & 58.2 & \multirow{2}{*}{51.5} \\
       & & 2 & 91.4 & 89.5 & 95.2 & 98.9 & 77.5 & 89.6 & 52.6 & 93.6 & 51.0 &  \\
      \addlinespace[2pt]
      \multirow{2}{*}{SimScale} & \multirow{2}{*}{CVPR 26} & 1 & 99.6 & 99.1 & 99.9 & 100.0 & 69.6 & 99.6 & 95.8 & 95.6 & 28.4 & \multirow{2}{*}{53.3} \\
       & & 2 & 94.5 & 94.2 & 95.8 & 99.2 & 75.8 & 92.8 & 60.1 & 96.1 & 43.2 &  \\
      \addlinespace[2pt]
      \multirow{2}{*}{DrivoR} & \multirow{2}{*}{CVPR 26} & 1 & 98.9 & 97.3 & 99.2 & 99.6 & 77.7 & 99.1 & 95.3 & 97.6 & 68.4 & \multirow{2}{*}{52.3} \\
       & & 2 & 92.3 & 92.2 & 97.0 & 99.0 & 72.1 & 90.3 & 56.3 & 97.1 & 38.8 & \\
      \addlinespace[2pt]
      \multirow{2}{*}{DrivoR-Scale} & \multirow{2}{*}{CVPR 26} & 1 & 99.1 & 98.2 & 99.3 & 99.8 & 75.4 & 98.7 & 94.9 & 97.6 & 70.2 & \multirow{2}{*}{54.6} \\
       & & 2 & 92.3 & 91.6 & 97.3 & 99.1 & 75.7 & 90.6 & 56.1 & 98.4 & 44.7 &  \\
      \addlinespace[2pt]
      \multirow{2}{*}{DriveZero} & \multirow{2}{*}{arXiv 26.09} & 1 & 98.7 & 98.9 & 99.9 & 99.8 & 79.5 & 98.7 & 95.8 & 97.6 & 51.6 & \multirow{2}{*}{51.5} \\
       & & 2 & 90.5 & 89.2 & 94.6 & 98.6 & 81.2 & 88.3 & 57.2 & 94.5 & 41.3 & \\
      \addlinespace[2pt]
      \multirow{2}{*}{DriveZero-Scale} & \multirow{2}{*}{arXiv 26.09} & 1 & 98.7 & 99.3 & 100.0 & 99.3 & 76.7 & 98.4 & 95.3 & 96.9 & 44.4 & \multirow{2}{*}{\underline{56.8}} \\
       & & 2 & 94.4 & 95.4 & 97.7 & 98.4 & 76.1 & 92.5 & 60.0 & 95.7 & 44.0 &  \\
      \midrule
      \rowcolor{gray!12}
       & & 1 & 99.3 & 98.4 & 99.7 & 99.6 & 78.0 & 99.1 & 96.0 & 97.3 & 50.7 & \\
      \rowcolor{gray!12}
      \multirow{-2}{*}{\textbf{\methodname{}}} & \multirow{-2}{*}{--} & 2 & 94.2 & 93.5 & 95.9 & 99.3 & 79.8 & 92.1 & 59.8 & 94.8 & 41.4 & \multirow{-2}{*}{55.9} \\
      \rowcolor{gray!12}
       & & 1 & 99.8 & 98.0 & 99.9 & 99.8 & 77.0 & 99.3 & 96.2 & 97.3 & 59.6 & \\
      \rowcolor{gray!12}
      \multirow{-2}{*}{\textbf{\methodname{}-Pro}} & \multirow{-2}{*}{--} & 2 & 95.4 & 92.8 & 95.1 & 99.3 & 80.5 & 93.5 & 59.0 & 96.8 & 47.1 & \multirow{-2}{*}{\textbf{58.1}} \\
      \bottomrule
    }
  \endgroup
\end{table*}
\begin{table*}[t]
  \centering
  \small
  \caption{\textbf{Zero-shot closed-loop evaluation on HUGSIM.} 
    Best average scores are in \textbf{bold}, second-best are \underline{underlined}.
    E/M/H/X denote Easy/Medium/Hard/Extreme.
    \methodname{} reports preliminary results on 345 scenarios. \textbf{Visualizations are shown in the supplementary video.}}
  \label{tab:hugsim-closed-loop}
  \begingroup
  \setlength{\tabcolsep}{4pt}
  \renewcommand{\arraystretch}{1.08}
    \fpdfullwidthtabular{24}{lccccccccccc}{
      \toprule
      \multirow{2}{*}{\textbf{Method}} & \multirow{2}{*}{\textbf{Venue}}
      & \multicolumn{5}{c}{\textbf{RC}$\uparrow$}
      & \multicolumn{5}{c}{\textbf{HD-Score}$\uparrow$} \\
      \cmidrule(lr){3-7}\cmidrule(lr){8-12}
      & & E & M & H & X & Avg. & E & M & H & X & Avg. \\
      \midrule
      VAD & ICCV 23 & 51.3 & 31.1 & 25.3 & 26.5 & 31.4 & 36.3 & 9.5 & 8.0 & 11.5 & 13.4 \\
      LTF & TPAMI 22 & 67.8 & 35.1 & 26.2 & 40.5 & 38.9 & 58.9 & 18.0 & 9.8 & 25.9 & 23.7 \\
      UniAD & CVPR 23 & 78.4 & 60.5 & 33.6 & 17.8 & 45.9 & 64.9 & 45.8 & 20.6 & 6.6 & 32.7 \\
      DrivoR & CVPR 26 & 80.9 & 50.5 & 33.8 & 47.1 & 49.8 & 73.3 & 34.6 & 18.8 & 32.5 & 35.7 \\
      GigaPixel & arXiv 26.06 & 78.6 & 60.7 & 32.7 & 35.4 & 50.1 & 67.4 & 51.9 & 19.1 & 21.6 & 38.5 \\
      DriveZero & arXiv 26.09 & 85.5 & 63.7 & 31.7 & 43.3 & \underline{53.9} & 78.5 & 47.2 & 16.0 & 28.7 & \underline{39.4} \\
      \midrule
      \rowcolor{gray!12}
      \textbf{\methodname{}} & - & 86.2 & 62.8 & 38.2 & 43.3 & \textbf{55.3} & 78.5 & 57.0 & 25.2 & 28.1 & \textbf{45.1} \\
      \bottomrule
    }
  \endgroup
\end{table*}

\begin{table*}[t]
\centering
\small
\caption{\textbf{Ablations of \methodname{}'s architecture and distillation components on NAVSIM v1.} Teacher PDMS is reported as learned-scorer selection / PDM-Closed evaluator selection.}
\label{tab:model_components_grouped}
\setlength{\tabcolsep}{5pt}
\renewcommand{\arraystretch}{1.15}
\fpdfullwidthtabular{18}{cccccc|cc|r@{\,/\,}l}{
\toprule
\multirow{2}{*}{Model}
& \multirow{2}{*}{Traj. Dec.}
& \multirow{2}{*}{Ref. DiT}
& \multirow{2}{*}{Global Scene Enc.}
& \multirow{2}{*}{Future Token}
& \multirow{2}{*}{Scorer}
& \multicolumn{2}{c|}{Distillation}
& \multicolumn{2}{c}{\multirow{2}{*}{PDMS}} \\
\cmidrule(lr){7-8}
& & & & & & FPMT & SASC & \multicolumn{2}{c}{} \\
\midrule
\multirow{5}{*}{Teacher}
& \cmark & \xmark & \xmark & \xmark & \cmark & -- & -- & 94.0 & 96.7 \\
& \cmark & \xmark & \xmark & \cmark & \cmark & -- & -- & 94.8 & 98.8 \\
& \cmark & \cmark & \xmark & \cmark & \cmark & -- & -- & 95.1 & 99.2 \\
& \cmark & \cmark & \cmark & \cmark & \xmark & -- & -- & -- & 98.8 \\
& \cmark & \cmark & \cmark & \cmark & \cmark & -- & -- & 95.6 & 99.6 \\
\midrule
\multirow{5}{*}{Student}
& \cmark & \xmark & \xmark & \xmark & \cmark & \xmark & \xmark & \multicolumn{2}{c}{94.0} \\
& \cmark & \cmark & \xmark & \xmark & \cmark & \xmark & \xmark & \multicolumn{2}{c}{94.1} \\
& \cmark & \cmark & \xmark & \xmark & \cmark & \cmark & \xmark & \multicolumn{2}{c}{94.7} \\
& \cmark & \cmark & \xmark & \xmark & \cmark & \xmark & \cmark & \multicolumn{2}{c}{94.2} \\
& \cmark & \cmark & \xmark & \xmark & \cmark & \cmark & \cmark & \multicolumn{2}{c}{94.9} \\
\bottomrule
}
\end{table*}

\begin{table*}[t]
\centering
\small
\caption{\textbf{Ablations of \methodname{}'s teacher-target supervision on NAVSIM v1.}
}
\label{tab:target_ablation}

\begingroup
\setlength{\tabcolsep}{3pt}
\renewcommand{\arraystretch}{1.07}

\begin{minipage}[t]{0.4\linewidth}
\vspace{0pt}
\centering
\begin{tabular*}{\linewidth}
{@{\extracolsep{\fill}}ccc@{}}
\toprule
\multicolumn{3}{c}{\textbf{(a) Multi-target distillation}}\\
\midrule
\shortstack{Target $K$}
& $\bar{J}$
& PDMS$\uparrow$\\
\midrule
1 & 1.0 & 94.6 \\
2 & 1.9 & 94.7 \\
\textbf{4 (default)} & 3.1 & 94.9\\
6 & 3.3 & 94.9\\
\bottomrule
\end{tabular*}
\end{minipage}\hfill%
\begin{minipage}[t]{0.565\linewidth}
\vspace{0pt}
\centering
\begin{tabular*}{\linewidth}
{@{\extracolsep{\fill}}lcc@{}}
\toprule
\multicolumn{3}{c}{\textbf{(b) Supervised trajectory selection and distillation policy}}\\
\midrule
Selection
& \shortstack{Distillation}
& \shortstack{Stu. PDMS$\uparrow$}\\
\midrule
Teacher Scorer & \textbf{w/o} Fut. (Self-Disl.)  & 94.1 \\
PDM-Closed     & \textbf{w/o} Fut. (Self-Disl.)  & 94.3 \\
Teacher Scorer & \textbf{w/} Fut. (FPD)         & 94.6 \\
PDM-Closed     & \textbf{w/} Fut. (FPD)        & 94.9 \\
\bottomrule
\end{tabular*}
\end{minipage}

\endgroup
\end{table*}

\subsection{Experimental Setup}
\label{sec:exp:setup}

\paragraph{Benchmarks and metrics.}
We evaluate on the \texttt{navtest} split of NAVSIM v1~\citep{dauner2024navsim} and report the Predictive Driver Model Score (PDMS) and its five components.
On the \texttt{navhard} split of NAVSIM v2~\citep{cao2025pseudosimulation}, we follow the official two-stage evaluation protocol and report the combined Extended PDMS (EPDMS), stage-wise scores, and component metrics. The metrics of NAVSIM v1 and v2 are detailed in Appendix~\ref{app:navsim-metrics}.
We additionally conduct a zero-shot closed-loop evaluation on HUGSIM~\citep{zhou2024hugsim} across Easy, Medium, Hard, and Extreme difficulty levels.
And we report the averages value of route completion (RC) and HD-Score for each level. The Appendix~\ref{sec:training_details} provides implementation details for training, inference, and dataset.


\subsection{Experimental Results}
\label{sec:exp:comparison}

\paragraph{Quantitative results.}
We introduce two model variants, \methodname{} and \methodname{}-Pro.
We train the base model on NAVSIM \texttt{trainval} (106K) and SimScale (134K), following~\citep{kirby2026drivor,xiaomi2026drivezero}.
At the same training-data scale, our base model achieves state-of-the-art performance among the compared methods (Table~\ref{tab:navsim-v1-leaderboard} and Table~\ref{tab:navsim-v2-leaderboard}). 
We attribute these gains to multiple trajectory supervision and more reliable candidate selection.
The future-privileged teacher provides high-quality scene-adaptive alternatives, extending supervision beyond the recorded human trajectory.
Safety-Aware Score Correction improves safety assessment, helping the student select safe trajectories more reliably.
As shown in Fig.~\ref{fig:exp:qual1}, our method achieves higher success rates in complex scenarios with modest computational overhead.
Moreover, zero-shot closed-loop results on HUGSIM (Table~\ref{tab:hugsim-closed-loop}) further support our model's generalization to closed-loop driving.
Motivated by the benefits of data augmentation in DriveZero-Scale~\citep{xiaomi2026drivezero}, we further train \methodname{}-Pro with additional generated data (158.8K, details in Appendix~\ref{sec:data_scaling}) to evaluate the benefits of expanding the training distribution.


\paragraph{Qualitative analysis.}
Fig.~\ref{fig:exp:qual1} compares the planned trajectory of the open-source method. In row~(a), every baseline leaves the road and hits a vehicle parked off the road; \methodname{} stays on the road and passes it. In row~(b), every baseline is on course to collide with the vehicle. The dashed paths show where each car would be if it kept its current speed. For the baselines, these land on the vehicle on the right;
\methodname{} would pass quickly. In row~(c), no method fails any metric, and the difference is progress: the baselines stop partway along the human route; \methodname{} follows it to the end. Safety does not cost progress. Additional visualizations are presented in Appendix~\ref{sec:exp_details}.

\paragraph{Ablation studies.}
We conduct all ablations using our base FPD-Drive.
Component ablations in Table~\ref{tab:model_components_grouped} examine the teacher and student designs.
For each teacher, the two PDMS values correspond to selection by the learned scorer and the PDM-Closed evaluator, respectively.
PDM-Closed selects higher-quality trajectories under the evaluation metrics, supporting its use for teacher-target selection.
However, we still train the scorer jointly with the teacher because gradients from metric prediction help improve scene representations for trajectory generation.
Future tokens and the global scene encoder improve the teacher's trajectory prediction, while residual DiT refinement benefits both the teacher and student, supporting our backbone design.
The student performs best when Future-Privileged Multi-Target Distillation and Safety-Aware Score Correction are combined, suggesting complementary contributions to trajectory generation and selection.

The ablation on multi-target distillation in Table~\ref{tab:target_ablation}(a) examines the effect of target budget.
Here, $K$ denotes the maximum target budget per scene, and $\bar{J}$ denotes the average number of targets retained after quality and diversity filtering.
Using multiple targets improves student performance, with diminishing gains at larger target budgets.
The ablation on supervised trajectory selection in Table~\ref{tab:target_ablation}(b) shows that selecting higher-quality targets with PDM-Closed improves student performance compared with learned-scorer selection.
Under both selection criteria, distillation from a future-privileged teacher outperforms self-distillation without future information.
Additional ablations on future-token injection locations, token-level distillation policy, distillation target sources and number of DiT denoising steps are provided in Appendix~\ref{sec:more_ablation}.


\section{Conclusion}

We presented \methodname{}, a framework that uses recorded futures to expand what a driving policy can learn from each demonstration.
A future-informed teacher provides multiple alternatives, which are evaluated and filtered into trajectory targets for the student.
Targeted score correction helps the student distinguish unsafe candidates from nearby safe alternatives and recognize safe options it would otherwise undervalue.
\methodname{} achieves state-of-the-art performance on NAVSIM v1 and v2, while zero-shot closed-loop evaluation on HUGSIM further demonstrates its generalization to closed-loop driving.
Limitations and future directions are discussed in Appendix~\ref{sec:limitations}.




\bibliography{iclr2027_conference}
\bibliographystyle{iclr2027_conference}


\clearpage
\appendix
\section*{\textbf{Appendix}}
\section{Training and Inference Details}
\label{sec:training_details}

Training consists of three stages.
In Stage~1, we train the coarse-to-fine planner described in Sec.~\ref{sec:method:planner}, which serves as the common backbone for the teacher and student, for 30 epochs, starting with a learning rate of $2\times10^{-4}$.
In Stage~2, we initialize the teacher from the Stage~1 model and train it for 10 epochs with privileged future observations, starting with a learning rate of $2\times10^{-4}$.
In Stage~3, we freeze the teacher and train the student, initialized from the Stage~1 model, using Future-Privileged Multi-Target Distillation and Safety-Aware Score Correction for 10 epochs, starting with a learning rate of $2\times10^{-5}$.
We conducted all training on 8$\times$H800 GPUs with a batch size of 16, using the AdamW optimizer and a cosine learning rate decay strategy with linear warmup. We adopt gradient clipping with global norm $1.0$, and the training precision is \textit{BF16}. Table~\ref{tab:student-config} details the model configuration.

\begin{figure}[t]
    \centering
    \includegraphics[width=0.9\linewidth]{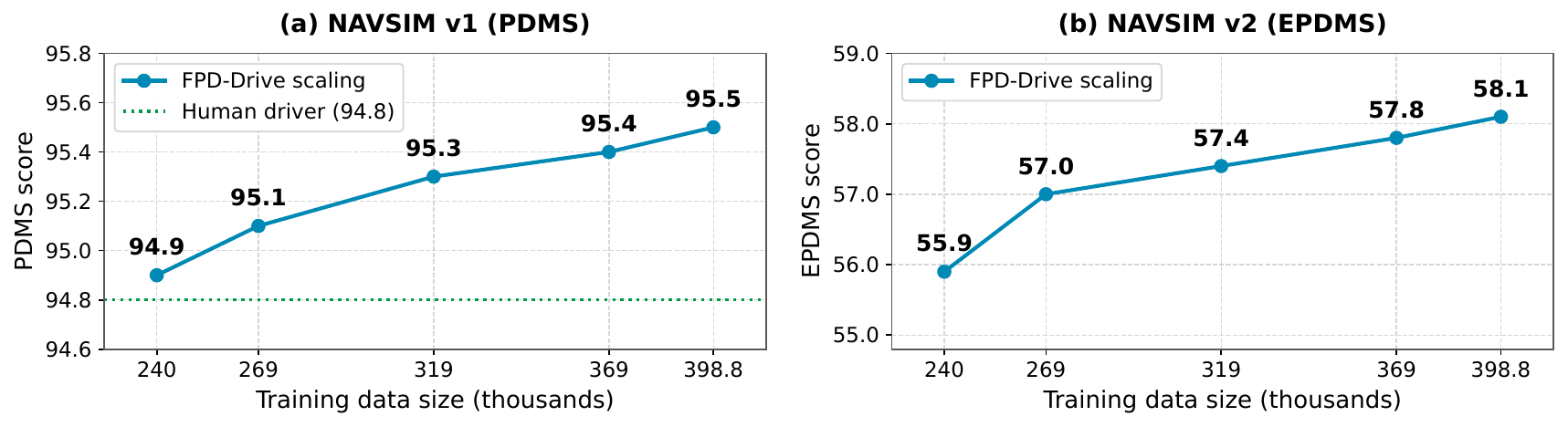}
    \caption{\textbf{Training-data scaling of \methodname{}.}}
    \label{fig:data_scaling}
\end{figure}

Following the training-data setup of DrivoR~\citep{kirby2026drivor}, we train our base model, \methodname{}, on approximately 106K NAVSIM \texttt{trainval} scenes and 134K SimScale scenes.
NAVSIM~\citep{dauner2024navsim} provides real-world driving logs derived from nuPlan, including multi-view images and human driving trajectories; SimScale~\citep{tian2026simscale} augments recorded scenes through reactive simulation, providing rendered multi-view observations and pseudo-expert trajectories.
We further train \methodname{}-Pro with additional controllably generated driving scenarios to broaden behavioral coverage, with generation details and data composition in Sec.~\ref{sec:data_scaling} and Table~\ref{tab:augmentation_budget}, and the resulting scaling trends in Fig.~\ref{fig:data_scaling}.

At inference, the model takes four camera views from the current frame, the current ego state, and a navigation command.
The refiner starts from Gaussian residual noise and runs 8 DDIM steps~\citep{song2021ddim} before adding the residual to each of the 64 trajectory proposals.
The scorer then selects one refined trajectory for execution.
The output is an $8\times3$ sequence of future ego poses $(x_h,y_h,\psi_h)$ at 0.5-second intervals, covering the next 4 seconds.
All poses are expressed in the current ego coordinate frame, with positions $(x_h,y_h)$ in meters and heading $\psi_h$ in radians.

\begin{table}[H]
  \centering
  \small
  \caption{\textbf{Model configuration.}}
  \label{tab:student-config}
  \begingroup
  \setlength{\tabcolsep}{5pt}
  \renewcommand{\arraystretch}{1.08}
    \begin{tabular*}{\textwidth}{@{\extracolsep{\fill}}p{0.38\textwidth}p{0.56\textwidth}@{}}
      \toprule
      \textbf{Component} & \textbf{Student configuration} \\
      \midrule
      Visual token dim & $4 \times 16 \times 256$ \\
      Visual backbone & DINOv2 (ViT-S/14 vision transformer), frozen \\
      Visual backbone adaptation & LoRA, rank 32 \\
      Cameras & Front, left, right, and rear \\
      Input resolution & $1148\times672$ \\
      Input history & Current frame only \\
      Ego status & 11-dim current-frame token \\
      Continuous action & Ego-relative $(x, y, \mathrm{yaw})$ poses \\
      Prediction horizon & 8 poses at 2~Hz (4~s) \\
      Trajectory proposals & 64 \\
      Trajectory tokens dim & $8 \times 256$ \\
      Attention heads & 8 \\
      Proposal decoder & 4 layers \\
      Scoring decoder & 4 layers, 5 MLP heads \\
      Residual refiner & 12-layer DiT, 8 heads \\
      Refinement process & DDIM, 8 denoise steps \\
      Model parameters &   123.3M  \\
      \bottomrule
    \end{tabular*}
  \endgroup
\end{table}

\section{Data Scaling}
\label{sec:data_scaling}
Common driving patterns dominate the scenario distribution, leaving
uncommon traffic conditions and safety-critical events
underrepresented. This imbalance limits the diversity of future
behaviors and responses that planners can learn from recorded
demonstrations.

We combined approximately 106K NAVSIM scenes and 134K SimScale scenes
as the original training data. To expand coverage of underrepresented
turning behaviors, we allocated approximately 60.6K generated scenes
to left commands, 96.4K to right commands, and 1.8K to straight commands.

\begin{figure*}[!th]
    \centering
    \includegraphics[width=\linewidth]{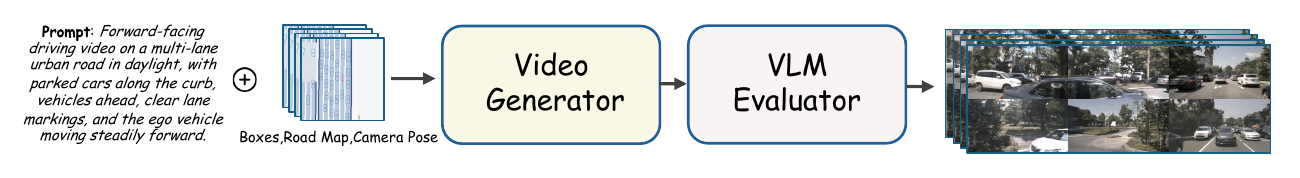}
    \caption{
    \textbf{Overview of the data generation pipeline.}
    }
    \label{fig:data_generation_overview}
\end{figure*}

Fig.~\ref{fig:data_generation_overview} illustrates our data generation
pipeline. We used controllable driving video
generation~\citep{chen2026vistagenconsistentdrivingvideo,gao2025magicdrivev2highresolutionlongvideo},
conditioned on text prompts, bounding-box layouts, road maps, and
camera poses. Manually specified layouts and object identity
descriptions control the objects, spatial relationships, and
interactions in each scenario which can be used to calculate metric labels. A vision-language model evaluator~\cite{bai2025qwen3vl} assesses the generated videos before their inclusion
in the augmentation pool.
Table~\ref{tab:augmentation_budget} summarizes the marginal quotas
applied within this pool. These generated samples provide
complementary supervision intended to broaden the training
distribution and support reasoning about diverse future outcomes. The generated data was used only for \textbf{\methodname{}-Pro},
while \textbf{\methodname{}} used only the public datasets.

\begin{table}[h]
    \centering
    \small
    \setlength{\tabcolsep}{4pt}
    \renewcommand{\arraystretch}{1.1}
    \caption{\textbf{Data allocation used for the 158.8K generated scenes.}}
    \label{tab:augmentation_budget}
    \begin{tabular*}{\textwidth}{@{\extracolsep{\fill}}lrr@{\hspace{1.5em}}lrr@{}}
        \toprule
        \multicolumn{3}{c}{\textit{Navigation command group}} & \multicolumn{3}{c}{\textit{Speed group}} \\
        \cmidrule(lr){1-3}\cmidrule(lr){4-6}
        Condition & \shortstack{Original\\data (\%)} & \shortstack{Generated\\data (K)}
        & Condition & \shortstack{Original\\data (\%)} & \shortstack{Generated\\data (K)} \\
        \midrule
        Left     & 25.36 & 60.6 & 2--5\,m/s    & 32.40 & 55.9 \\
        Right    & 11.46 & 96.4 & 5--10\,m/s   & 35.89 & 46.4 \\
        Straight & 63.18 &  1.8 & Other speeds & 31.71 & 56.5 \\
        \bottomrule
    \end{tabular*}
\end{table}

\section{NAVSIM Evaluation Metrics}
\label{app:navsim-metrics}

NAVSIM evaluates planning in a pseudo-closed-loop setting.
An LQR controller tracks the predicted 4~s trajectory using a kinematic
bicycle model at 10~Hz. The resulting ego states are evaluated against
the map, route, and surrounding traffic.
Each sub-metric takes a value in $[0,1]$ and is reported as a percentage.
The composite score combines multiplicative factors with a fixed weighted
average of the remaining metrics.
NAVSIM v1 and v2 differ in both their metric sets and scenario organization.

NAVSIM v1 (\texttt{navtest}) uses two multiplicative factors:
no at-fault collisions (NC) and drivable-area compliance (DAC).
NC considers only collisions for which the ego vehicle is responsible,
while DAC checks whether the vehicle footprint remains within the
drivable area. The weighted component includes ego progress (EP),
time-to-collision within bounds (TTC), and comfort (C).
EP measures progress along the centerline relative to the PDM-Closed
reference. TTC checks whether the time to collision with any other
vehicle remains above a fixed threshold. C checks whether the simulated
trajectory satisfies bounds on acceleration, jerk, and yaw rate.
The composite score is:
\begin{equation}
\mathrm{PDMS} = \mathrm{NC}\cdot\mathrm{DAC}
\times\frac{5\,\mathrm{EP} + 5\,\mathrm{TTC} + 2\,\mathrm{C}}{12}.
\end{equation}

NAVSIM v2 (\texttt{navhard\_two\_stage}) adds two multiplicative factors:
driving-direction compliance (DDC) and traffic-light compliance (TLC).
DDC penalizes travel against the lane direction within a $1$~s window.
Its score is $1$ for wrong-way travel below $2$~m, $0.5$ for travel
between $2$~m and $6$~m, and $0$ for travel beyond $6$~m.
TLC checks whether the ego vehicle crosses a stop line at a red light.

The weighted component additionally includes lane keeping (LK), which
checks lateral deviation from the centerline using a $0.5$~m threshold
over a $2$~s window. The original comfort metric is replaced by history
comfort (HC) and extended comfort (EC).
HC evaluates the simulated trajectory together with the recorded past
to assess comfort at their junction.
EC concatenates trajectories predicted at two adjacent frames of the
same scene and applies the comfort bounds to the combined sequence.
This penalizes inconsistencies between successive planning steps.
The extended composite score is:
\begin{equation}
\mathrm{EPDMS} = \mathrm{NC}\cdot\mathrm{DAC}\cdot\mathrm{DDC}\cdot\mathrm{TLC}
\times\frac{5\,\mathrm{EP} + 5\,\mathrm{TTC} + 2\,\mathrm{LK}
+ 2\,\mathrm{HC} + 2\,\mathrm{EC}}{16}.
\end{equation}

NAVSIM v2 also changes how it organizes scenarios and aggregates scores.
Its $5{,}912$ scenarios form $225$ groups, each evaluated from two start
frames separated by $0.5$~s.
The first stage evaluates the original recorded frames.
The second stage evaluates synthetic continuations in which surrounding
traffic responds to the planned ego behavior.
For each start frame, scores are averaged separately within each stage,
and the two stage averages are multiplied.
The resulting scores for the two start frames are then averaged to
obtain the group score. The final EPDMS is the mean across all groups.

\begin{figure*}[!ht]
    \centering
    \includegraphics[width=1.0\linewidth]{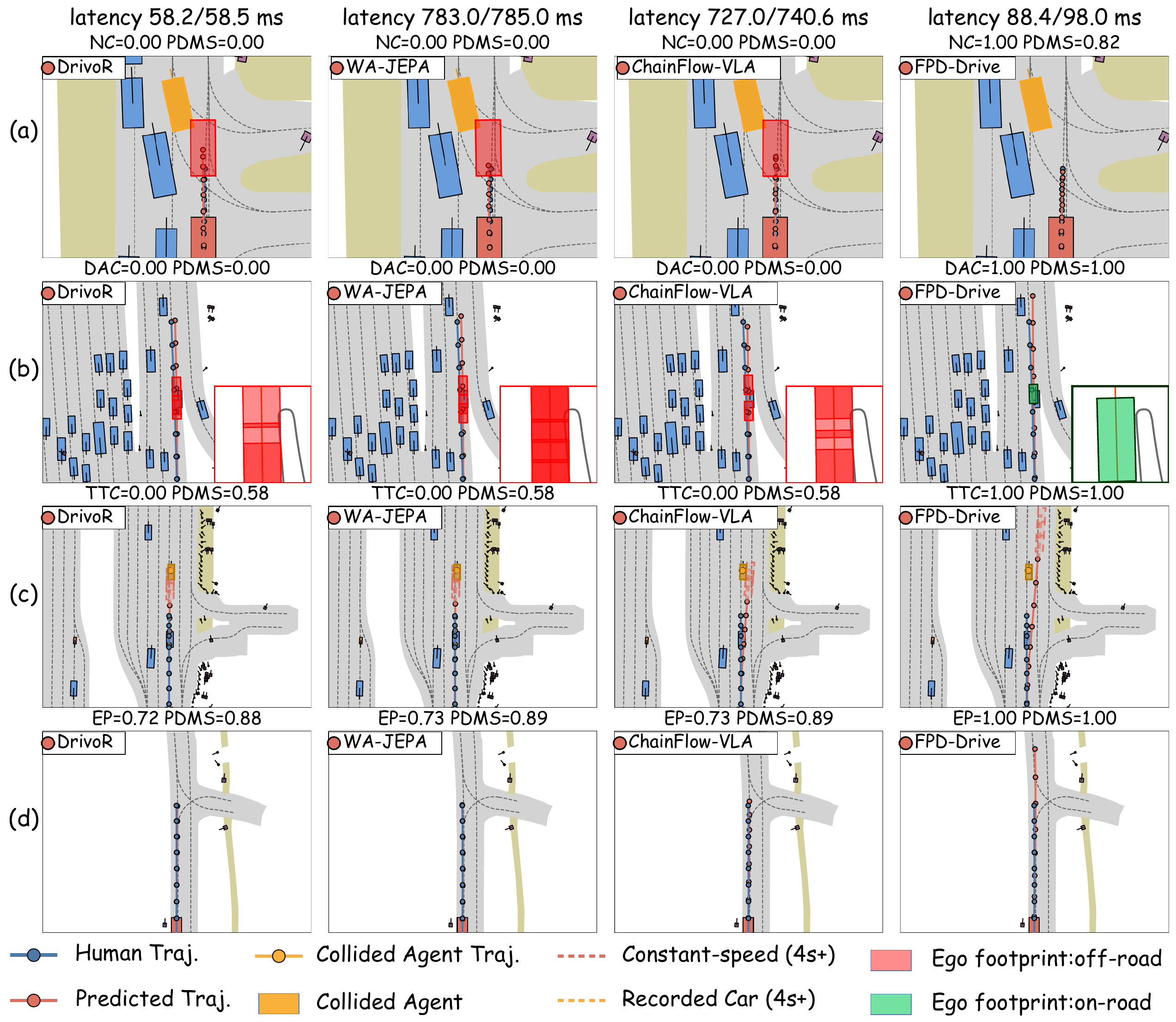}
    \caption{\textbf{Additional NAVSIM \texttt{navtest} scenarios.}}
    \label{fig:exp:qual2}
\end{figure*}

\section{Experiment Details}
\label{sec:exp_details}


\paragraph{Qualitative analysis.}
Fig.~\ref{fig:exp:qual2} shows additional examples of planned trajectories from open-source methods on single H800 GPU.
Row~(a) is an outright crash: every baseline drives into the car ahead, while \methodname{} keeps clear of it. Row~(b) is about road boundaries: at a corner, every baseline cuts the curve and leaves the drivable area, which zeroes DAC; \methodname{} stays on the road. Row~(c) repeats the TTC pattern of the main paper in a new scene --- the dashed constant-speed paths of the baselines again land on the car ahead; that of \methodname{} stays well clear of it. Row~(d) shows the progress side: with every metric satisfied by every method, \methodname{} travels 37\,m while the baselines travel 25--26\,m, close to the recorded human route of 25\,m.

\begin{table}[!ht]
\centering
\small
\setlength{\tabcolsep}{6pt}
\renewcommand{\arraystretch}{1.15}
\caption{\textbf{Ablation study on future-token injection in the privileged teacher. }}
\label{tab:future_token_injection}
\begin{tabular*}{\textwidth}{@{\extracolsep{\fill}}c|ccc|c@{}}
\toprule
\textbf{Setting} 
& \textbf{Traj. Dec.} 
& \textbf{Ref. DiT} 
& \textbf{Scorer} 
& \textbf{PDMS}$\uparrow$ \\
\midrule
Causal baseline (No Future Token)
& $\times$ & $\times$ & $\times$ & 94.13 \\
Decoder only
& $\checkmark$ & $\times$ & $\times$ & 94.26 \\
DiT only
& $\times$ & $\checkmark$ & $\times$ & 95.29 \\
Scorer only
& $\times$ & $\times$ & $\checkmark$ & 94.79 \\
Decoder + Scorer
& $\checkmark$ & $\times$ & $\checkmark$ & 94.89 \\
Decoder + DiT
& $\checkmark$ & $\checkmark$ & $\times$ & 95.37 \\
Decoder + DiT + Scorer
& $\checkmark$ & $\checkmark$ & $\checkmark$ & 95.41 \\
\textbf{Ref. DiT + Scorer (Ours Teacher)}
& $\times$ & $\checkmark$ & $\checkmark$ & \textbf{95.57} \\
\bottomrule
\end{tabular*}
\end{table}

\section{More Ablation Studies}
\label{sec:more_ablation}

\paragraph{Future-token injection in the teacher.}
We investigate where to introduce future tokens in the privileged teacher.
Starting from a causal teacher without future access, we evaluate three possible
injection locations: the trajectory decoder, residual DiT refiner, and scorer.
Future conditioning improves the teacher across different configurations.
Interestingly, injecting future tokens into all modules does not provide further
benefits. The Ref. DiT + Scorer configuration achieves the best performance,
suggesting that future information is most valuable for trajectory refinement
and candidate evaluation rather than initial trajectory generation. The quantitative results are shown in Table~\ref{tab:future_token_injection}.

\paragraph{Token-level versus multi-target distillation.}
Inspired by privileged policy distillation in EvoDriveVLA~\citep{cao2026evodrivevla}, we compare token-level distillation with our Future-Privileged Multi-Target Distillation.
We adapt their token-distribution matching idea to our diffusion planner by applying a token-level KL divergence between teacher and student distributions during DiT denoising.
The teacher has access to recorded future observations, whereas the student uses only current observations.
The token-level variant underperforms multi-target distillation, as summarized in Table~\ref{tab:distillation_granularity}.
This suggests that matching token distributions during denoising is less effective in our setting than explicitly supervising multiple complete trajectories selected for planning quality and geometric diversity.
Future-Privileged Multi-Target Distillation assigns these alternatives to distinct student proposals, directly encouraging coverage of multiple valid behaviors in the same scene.

\paragraph{Ablation of distillation target sources.}
We ablate the candidate-trajectory source for multi-target supervision.
Inspired by CLOVER ~\cite{ang2026clover}, we construct 64 predefined trajectory vocabularies and compare them with 64 trajectories generated by our future-privileged teacher for each scene.
For top-$K$ supervision, both sources use the same target budget $K=4$, PDM-Closed evaluation, and safety and diversity criteria described in Sec.~\ref{sec:method:multitarget}.
Table~\ref{tab:distillation_target_source} shows that teacher-generated candidates provide more effective supervision, supporting the value of learning scene-adaptive alternatives beyond manually specified motion patterns.

\begin{table}[!ht]
    \centering
    \small
    \setlength{\tabcolsep}{8pt}
    \renewcommand{\arraystretch}{1.1}
    \caption{\textbf{Token-level versus multi-target distillation on NAVSIM v1.}
    }
    \label{tab:distillation_granularity}
    \begin{tabular*}{\textwidth}{@{\extracolsep{\fill}}lc@{}}
        \toprule
        Distillation strategy & PDMS$\uparrow$ \\
        \midrule
        EvoDrive-VLA & 85.3 \\
        Ours w/ Token-level KL & 94.5 \\
        FPD-Drive & 94.9 \\
        \bottomrule
    \end{tabular*}
\end{table}

\begin{table}[!ht]
    \centering
    \small
    \setlength{\tabcolsep}{8pt}
    \renewcommand{\arraystretch}{1.1}
    \caption{\textbf{Ablation of distillation target sources on NAVSIM v1.}
    Both variants supervise our student using up to $K=4$ targets selected from 64 candidate trajectories per scene.}
    \label{tab:distillation_target_source}
    \begin{tabular*}{\textwidth}{@{\extracolsep{\fill}}lcc@{}}
        \toprule
        Candidate source & Candidates & Stu. PDMS$\uparrow$ \\
        \midrule
        Predefined trajectory vocabularies & 64 & 94.2 \\
        Future-privileged teacher (ours) & 64 & \textbf{94.9} \\
        \bottomrule
    \end{tabular*}
\end{table}

\paragraph{Components of Safety-Aware Score Correction.}
We ablate the two regularization terms $\mathcal{L}_{\mathrm{neg}}$ and $\mathcal{L}_{\mathrm{pos}}$ in ~\eqref{eq:method:negative} and~\eqref{eq:method:positive}.
We report PDMS and the percentage of unsafe cases in the test set, where a case is unsafe if the selected trajectory has a zero value of NC or DAC.
As shown in Table~\ref{tab:safety_correction_losses}, combining the two terms yields the lowest unsafe-case rate.

\begin{table}[!ht]
    \centering
    \small
    \setlength{\tabcolsep}{8pt}
    \renewcommand{\arraystretch}{1.1}
    \caption{\textbf{Ablation of the Components of SASC.}}
    \label{tab:safety_correction_losses}
    \begin{tabular*}{\textwidth}{@{\extracolsep{\fill}}cccc@{}}
        \toprule
        $\mathcal{L}_{\mathrm{neg}}$ & $\mathcal{L}_{\mathrm{pos}}$ & Unsafe cases (\%)$\downarrow$ & PDMS$\uparrow$ \\
        \midrule
        \xmark & \xmark & 1.91 & 94.7 \\
        \cmark & \xmark & 1.77 & 94.8 \\
        \xmark & \cmark & 1.89 & 94.7 \\
        \cmark & \cmark & \textbf{1.35} & \textbf{94.9} \\
        \bottomrule
    \end{tabular*}
\end{table}

\paragraph{Number of DiT denoising steps.}
We study \methodname{} 's sensitivity to the number of DDIM denoising steps~\citep{song2021ddim}, using 2, 4, 6, 8, and 12 steps with the same trained model.
This ablation examines how much iterative refinement is needed for effective planning and whether additional denoising steps justify their inference cost.
Table~\ref{tab:denoising_steps} summarizes the settings.

\begin{table}[!ht]
    \centering
    \small
    \setlength{\tabcolsep}{12pt}
    \renewcommand{\arraystretch}{1.1}
    \caption{\textbf{Ablation of DiT denoising steps on NAVSIM v1.}
    }
    \label{tab:denoising_steps}
    \begin{tabular*}{\textwidth}{@{\extracolsep{\fill}}lccccc@{}}
        \toprule
        Denoising steps & 2 & 4 & 6 & 8 & 12 \\
        \midrule
        PDMS$\uparrow$ & 94.7 & 94.7 & 94.8 & \textbf{94.9} & 94.9 \\
        \bottomrule
    \end{tabular*}
\end{table}

\section{Limitations}
\label{sec:limitations}

\methodname{} still incurs inference overhead from iterative DiT sampling.
Future work will explore single-step denoising and refinement, motivated by MeanFuser~\citep {wang2026meanfuserfastonestepmultimodal} 's efficient one-step trajectory generation.
Further evaluation is needed to assess robustness during longer-horizon interactions.
We have not evaluated the model under extreme weather conditions, such as heavy rain and snow, or deployed it on real vehicles, so we leave these settings for future validation.

\end{document}